\documentclass{article}

\usepackage[final]{corl_2026} % Uncomment for the camera-ready ``final'' version.
\usepackage{graphicx}
\usepackage{booktabs}
\usepackage{multirow}
\usepackage{xcolor}
\usepackage{wrapfig}
\usepackage{amsmath}
\usepackage{amssymb}

\usepackage{hyperref}
\usepackage{enumitem}

\begin{document}

% \title{KPGrasp: Learning High-quality Dexterous Grasps via Scalable Keypoint Flow Matching} 

% \title{KPGrasp: Heuristic-Free Dexterous Grasp Generation via Scalable Keypoint Flow Matching}

% \title{KPGrasp: Keypoint Learning is All You Need \\ for Dexterous Grasp Generation}

\title{KPGrasp: Scalable Keypoint Flow Matching \\ for Dexterous Grasp Generation}


\author{
  Yuansen Huang$^{1,2,*}$ \quad
  Jiayi Chen$^{1,2,*}$ \quad
  Haoran Liu$^{1,2}$ \quad
  Yubin Ke$^{1,2}$ \quad
  Bing Han$^{2,3}$ \\
  \textbf{Jiangran Lyu}$^{1,2}$ \quad
  \textbf{Mi Yan}$^{1,2}$ \quad
  \textbf{Li Yi}$^{2,4}$ \quad
  \textbf{He Wang}$^{1,2,\dagger}$ \\[1mm]
  $^{1}$Peking University \quad
  $^{2}$Galbot \quad
  $^{3}$Xi'an Jiaotong University \quad
  $^{4}$Tsinghua University \\[0.5mm]
  $^{*}$Equal contribution \quad
  $^{\dagger}$Corresponding author
}
% \author{
%   Jane E.~Doe\\
%   Department of Electrical Engineering and Computer Sciences\\
%   University of California Berkeley 
%   United States\\
%   \texttt{janedoe@berkeley.edu} \\
  %% examples of more authors
  %% \And
  %% Coauthor \\
  %% Affiliation \\
  %% Address \\
  %% \texttt{email} \\
  %% \AND
  %% Coauthor \\
  %% Affiliation \\
  %% Address \\
  %% \texttt{email} \\
  %% \And
  %% Coauthor \\
  %% Affiliation \\
  %% Address \\
  %% \texttt{email} \\
  %% \And
  %% Coauthor \\
  %% Affiliation \\
  %% Address \\
  %% \texttt{email} \\

\maketitle

\newcommand{\ys}[1]{\textcolor{blue}{TODO(Yuansen):#1}}
\newcommand{\jy}[1]{\textcolor{red}{TODO(Jiayi):#1}}
\newcommand{\cm}[1]{\textcolor{green}{Modification:#1}}

\begin{abstract}
Generating high-quality dexterous grasps remains challenging for learning-based methods, which often depend on carefully tuned contact losses or costly contact-based test-time refinement. We present KPGrasp, a flow-matching framework that learns dexterous grasp priors from large-scale data rather than relying on contact losses or contact-based test-time refinement. KPGrasp couples an all-Euclidean 3D hand-keypoint parameterization with a simple yet scalable Transformer flow model. The parameterization avoids the drawbacks of the conventional mixed $SE(3)$ pose and joint-angle output space, expresses grasps in the same frame as the object point cloud, and thus enables native spatial reasoning; the Transformer flow model is trained with only the standard flow-matching loss and scales effectively with data, model capacity, and batch size. Experiments demonstrate state-of-the-art performance on two simulation benchmarks. On the Dexonomy benchmark, it reaches a 76.3\% grasp success rate, improving over the strongest directly comparable baseline by 47.4\% while reducing penetration depth to 2.4 mm. The same model also achieves the best average performance on the DexGrasp Anything benchmark without fine-tuning. For batched inference, KPGrasp requires only 0.032 s per grasp. Finally, real-world experiments on 20 diverse objects demonstrate that the pipeline can be deployed in a real-world setup. \textbf{Project Page: \url{https://yuansen-huang.github.io/KPGrasp-page/}}
\end{abstract}

\begin{figure}[h]
    \centering
    \includegraphics[width=0.99\linewidth]{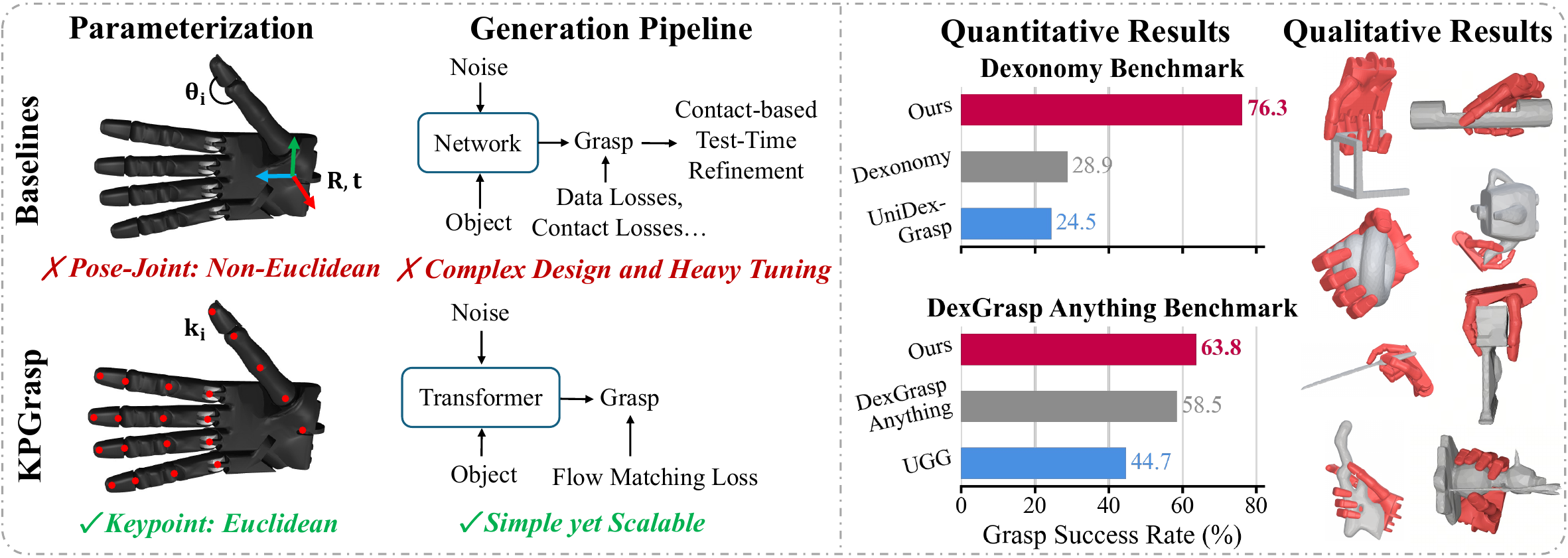}
    \caption{\textbf{Overview}. Unlike prior methods that use pose-joint parameterizations and rely on auxiliary contact losses or contact-based test-time refinement, KPGrasp parametrizes grasps with Euclidean hand keypoints and learns a simple yet scalable Transformer-based flow-matching model. KPGrasp achieves state-of-the-art performance on two simulation benchmarks and generates high-quality, diverse dexterous grasps.}
    \label{fig: teaser}
    \vspace{-3mm}
\end{figure}

\section{Introduction}
\label{sec:intro}
\vspace{-3mm}
Dexterous grasping is a core prerequisite for complex robotic manipulation, but generating reliable hand configurations remains difficult because valid grasps depend on both object geometry and the high-dimensional kinematics of the hand. Traditional analytic methods~\citep{miller2004graspit, ciocarlie2007dexterous, liu2021synthesizing, li2023frogger, chen2024springgrasp, chen2024task, chen2025bodex, chen2025dexonomy} address this problem by optimizing force-closure~\citep{ferrari1992planning} and related physics-based metrics to obtain physically stable grasps. Although these methods typically require accurate object meshes and are too expensive for fast online inference, they have become powerful engines for large-scale offline data generation~\citep{wang2022dexgraspnet, zhang2024dexgraspnet, ye2025dex1b, he2025dexvlg}. Recent frameworks such as Dexonomy~\citep{chen2025dexonomy}, for example, have synthesized 9.5M diverse grasps spanning the full taxonomy of human grasp types~\citep{feix2015grasp}.

This progress suggests a natural learning paradigm: use analytic methods offline to distill hand-crafted physics priors into large grasp datasets, then train neural networks to model the resulting grasp distribution directly. In practice, however, many learning-based methods reintroduce explicit physics priors into the learning pipeline, including contact, penetration, or physics-aware objectives during training~\citep{jiang2021hand,zhu2021toward,xu2024dexterous,zhong2025dexgrasp}, contact-based refinement or adaptation at test time~\citep{jiang2021hand,xu2023unidexgrasp,xu2024dexterous,li2022gendexgrasp}, or auxiliary evaluators and discriminators~\citep{lu2024ugg,weng2024dexdiffuser}. These components inherit some drawbacks of analytic methods, such as brittle parameter tuning and slow inference, and can still yield grasps with large penetrations or poor stability. This raises a central question: \textit{Can large-scale, high-quality data implicitly provide sufficient grasp priors for learning, without requiring hand-crafted designs that explicitly inject physics priors into the learning objective or test-time generation process?}

We answer this question with KPGrasp, a dexterous grasp generator that learns grasp priors directly from large-scale offline synthesized grasp data, without explicit contact losses or contact-based test-time refinement. KPGrasp combines an all-Euclidean 3D hand-keypoint parameterization with a simple and scalable Transformer-based flow-matching model, achieving state-of-the-art performance on two simulation benchmarks.

\begin{figure}
    \centering
    \includegraphics[width=0.99\linewidth]{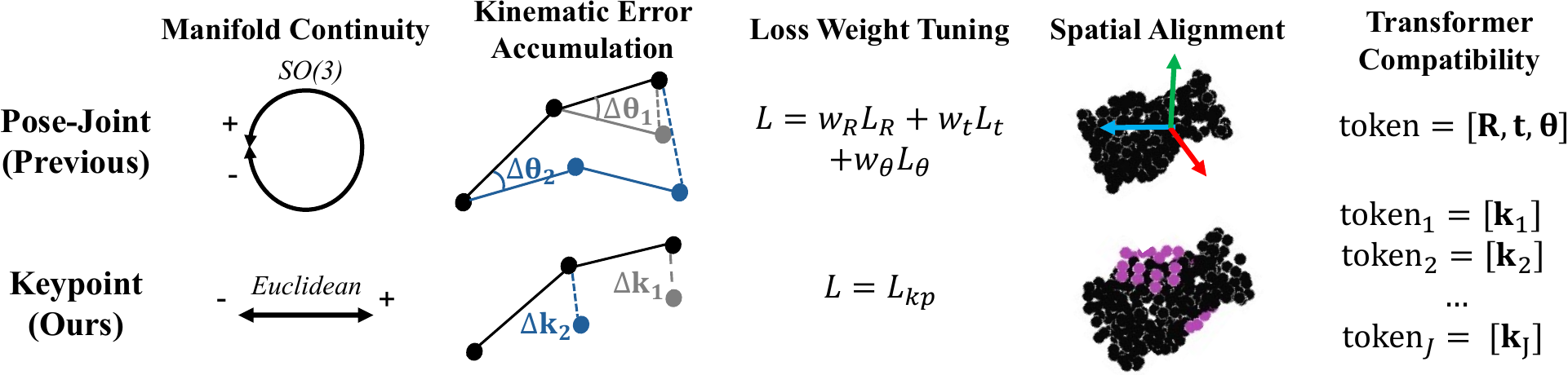}
    \caption{\textbf{Motivation for Keypoint Output Parameterization.} Traditional pose-joint outputs mix non-Euclidean wrist pose with joint angles, leading to SO(3) discontinuities~\citep{zhou2019continuity}, kinematic error accumulation, and loss balancing issues. Our keypoint outputs place the hand in Euclidean space, align naturally with object point clouds, and map cleanly to Transformer tokens.}
    \label{fig: motivation}
    \vspace{-3mm}
\end{figure}

The first key design choice is the grasp output parameterization. Most existing methods represent a grasp as a hybrid tuple consisting of a non-Euclidean $SE(3)$ wrist pose and Euclidean joint angles. As illustrated in Fig.~\ref{fig: motivation}, this pose-joint parameterization is not ideal for scalable generative learning: rotation prediction introduces manifold discontinuities~\citep{zhou2019continuity}, small wrist-pose errors can propagate along the kinematic chain, and mixed output spaces often require careful loss balancing.

KPGrasp instead predicts 3D hand joint keypoints directly. These keypoints lie in standard Euclidean space, align the hand output with the object point-cloud input, and allow the network to reason directly about spatial proximity during hand-object interaction. They also map naturally to Transformer architectures: each keypoint can be represented as a token, enabling attention layers to model dependencies across the hand structure.

Together, the keypoint parameterization and Transformer flow model play complementary roles. The former turns grasp generation into a clean Euclidean learning problem, while the latter shows that this space supports scalable generative modeling with only a standard flow-matching objective. 

Experiments demonstrate state-of-the-art performance on both the Dexonomy and DexGrasp Anything benchmarks. On Dexonomy, KPGrasp reaches a 76.3\% grasp success rate, improving over the strongest directly comparable baseline by 47.4 percentage points while reducing average penetration depth to 2.4 mm. On DexGrasp Anything, the same model achieves the best average success rate and lowest average penetration depth across five object sets without benchmark-specific fine-tuning. Ablations show that the keypoint parameterization improves grasp quality and that performance scales with dataset size, model capacity, and batch size. Without contact-based refinement, KPGrasp requires only 0.032 s per grasp during batched inference. Finally, real-world experiments on 20 diverse objects achieve an 83\% success rate.

In summary, our contributions are:
\begin{itemize}[nosep,leftmargin=*]
\item We propose KPGrasp, a dexterous grasp generator that learns grasp priors directly from large-scale data without contact losses or contact-based test-time refinement.
\item We introduce an all-Euclidean 3D hand-keypoint output parameterization to dexterous grasp generation that replaces the conventional hybrid $SE(3)$ pose and joint-angle representation.
\item We develop a Transformer-based flow model for keypoint grasp generation and show it scales with dataset size, model capacity, and batch size under the standard flow-matching objective.
\item We demonstrate state-of-the-art performance on the Dexonomy and DexGrasp Anything simulation benchmarks, efficient inference, and deployment in real-world grasping experiments.
\end{itemize}

\vspace{-2mm}
\section{Related Work}
\vspace{-2mm}

\subsection{Data-Driven Dexterous Grasp Generation}
\vspace{-2mm}

Early learning-based methods formulate grasp generation as a regression problem~\citep{liu2020deep, chen2022learning}, but deterministic predictors struggle to capture the multi-modal nature of valid grasp distributions. More recent approaches therefore adopt generative models, including conditional variational autoencoders~\citep{ye2025dex1b, jiang2021hand}, diffusion models~\citep{weng2024dexdiffuser, zhong2025dexgrasp}, normalizing-flow models~\citep{xu2023unidexgrasp, feng2024ffhflow}.
A common design direction is to compensate for the difficulty of learning hand-object interactions with task-specific losses, auxiliary discriminators, or contact-based test-time refinement. During training, many methods add contact or penetration penalties to the generative objective~\citep{jiang2021hand,zhu2021toward,xu2024dexterous,zhong2025dexgrasp}, making performance sensitive to loss-weight tuning. D(R,O) Grasp~\citep{wei2024mathcal} further introduces an explicit hand-object representation, but its pairwise structure incurs $O(N^2)$ memory cost. During inference, several methods refine generated grasps through optimization or test-time adaptation~\citep{jiang2021hand, li2022gendexgrasp, xu2023unidexgrasp, xu2024dexterous}, which improves feasibility but reintroduces the computational cost that learning-based methods seek to avoid. DexGrasp Anything~\citep{zhong2025dexgrasp} integrates physics-inspired guidance into the diffusion denoising process, but this still requires slow iterative computation. Other methods use auxiliary discriminators or evaluators to reject or refine low-quality grasps~\citep{weng2024dexdiffuser,lu2024ugg}, requiring additional training on imperfect grasp labels. In contrast, KPGrasp shows that a simple flow-matching objective and scalable Transformer architecture achieve state-of-the-art performance without contact losses, external discriminators, or contact-based test-time refinement.

\vspace{-2mm}
\subsection{Hand Parameterization}
\vspace{-2mm}

Most work parametrizes a grasp as a non-Euclidean $SE(3)$ wrist pose with Euclidean joint angles, introducing discontinuities, kinematic error propagation, and loss-balancing challenges. Prior work addresses these issues with specialized geometric modeling, such as $SE(3)$ flow matching~\citep{shi2025hograspflow}, $SE(3)$-equivariant networks~\citep{lim2024equigraspflow, zhu2025se}, or cascaded prediction that first estimates the root pose and then predicts joint angles~\citep{zhang2024dexgraspnet, chen2025dexonomy}.
An alternative is to represent the hand directly with 3D keypoints. Keypoint-based hand parameterizations are widely used in computer vision for hand pose estimation and tracking~\citep{cheng2021handfoldingnet, chen2023tracking, cheng2024handdiff}, where they provide a compact Euclidean description of articulated hand geometry. In robotics, recent work has begun to explore keypoints as policy outputs or unified action spaces for closed-loop manipulation~\citep{haldar2025point, goswami2025world, guzey2025dexterity}. However, their use as the output space for large-scale generative dexterous grasping remains under-explored. KPGrasp provides evidence that keypoint parameterization greatly improves the performance when trained on large grasp datasets.

\vspace{-2mm}
\section{Preliminary}
\vspace{-2mm}

\textbf{Flow Matching (FM).} Flow matching learns a time-dependent velocity field $\mathbf{v}$ that transports a simple prior $p_{\text{prior}}$ to the data distribution $p_{\text{data}}$~\citep{lipman2022flow}. We use the standard linear interpolation $\mathbf{x}_\tau=(1-\tau)\mathbf{x}_0+\tau\mathbf{x}_1$ for $\tau\in[0,1]$, where $\mathbf{x}_0\sim p_{\text{prior}}$ is Gaussian noise and $\mathbf{x}_1\sim p_{\text{data}}$ is a data sample. This path has conditional velocity $d\mathbf{x}_\tau/d\tau=\mathbf{x}_1-\mathbf{x}_0$, and FM trains $\mathbf{v}_\phi$ by minimizing:
\begin{equation}
\label{eq:loss}
\mathcal{L}_{\text{FM}} = \mathbb{E}_{\tau, \mathbf{x}_1 \sim p_{\text{data}}, \mathbf{x}_0 \sim p_{\text{prior}}} \left\| \mathbf{v}_\phi(\mathbf{x}_\tau, \tau) - (\mathbf{x}_1 - \mathbf{x}_0) \right\|_2^2
\end{equation}
At inference, samples are generated by solving the ODE forward from $\mathbf{x}_0 \sim p_{\text{prior}}$.

\textbf{Likelihood Computation.} The learned vector field also parameterizes a Continuous Normalizing Flow (CNF), so the log-density of a generated grasp can be accumulated during ODE integration by the instantaneous change-of-variables formula:
\begin{equation}
    \log p_{\text{data}}(\mathbf{x}_1) = \log p_{\text{prior}}(\mathbf{x}_0) - \int_0^1 \text{Tr}\left(\frac{\partial \mathbf{v}_\phi(\mathbf{x}_\tau, \tau)}{\partial \mathbf{x}_\tau}\right) d\tau
\end{equation}
The trace term is the divergence of the velocity field. Computing it exactly scales with the data dimension, so we follow standard CNF practice~\citep{grathwohl2018ffjord, song2020score, lipman2022flow} and use the Hutchinson estimator $\text{Tr}(\partial \mathbf{v}_\phi/\partial \mathbf{x}_\tau)=\mathbb{E}_{\epsilon}[\epsilon^\top(\partial \mathbf{v}_\phi/\partial \mathbf{x}_\tau)\epsilon]$ with a zero-mean unit-covariance probe vector $\epsilon$, which gives an unbiased $O(1)$ estimate per integration step. KPGrasp uses this tractable likelihood as a model-confidence score to rank generated grasp candidates, without the need of external discriminators.

\vspace{-3mm}
\section{Method}
\vspace{-2mm}

\subsection{Keypoint Grasp Parameterization}
\vspace{-2mm}

\label{sec:keypoint_grasp_parameterization}

\textbf{Notation.} A conventional dexterous grasp is represented as $[\mathbf{R}, \mathbf{t}, \boldsymbol{\theta}]$, where $\mathbf{R} \in SO(3)$ and $\mathbf{t} \in \mathbb{R}^{3}$ define the wrist pose, and $\boldsymbol{\theta} \in \mathbb{R}^{d_\theta}$ denotes the joint angles. KPGrasp instead predicts a set of 3D hand keypoints, $\mathbf{K}=[\mathbf{k}_1,\mathbf{k}_2,\ldots,\mathbf{k}_J]^\top \in \mathbb{R}^{J \times 3}$, in the same coordinate frame as the input object point cloud. Each $\mathbf{k}_j \in \mathbb{R}^{3}$ corresponds to a predefined point on the robot hand.

\textbf{Keypoint Definition.} As shown in Fig.~\ref{fig: teaser}, we use $J=21$ keypoints for the Shadow Hand. Keypoints are placed at the mechanical joints and fingertips, with five additional keypoints on the rigid palm. The palm keypoints consist of one palm-center point and four finger-root points. This particular palm-keypoint layout is not unique; any fixed set of reasonably spread points on the rigid palm can provide stable correspondences for wrist-pose registration.

\textbf{Parameterization Conversion.} Training labels $\mathbf{K}$ are obtained by applying forward kinematics to the pose-joint grasps $[\mathbf{R}, \mathbf{t}, \boldsymbol{\theta}]$ in the dataset. At deployment, the predicted keypoints are converted back to executable robot commands. The palm keypoints determine $(\mathbf{R}, \mathbf{t})$ through SVD-based rigid registration~\citep{umeyama1991least}, and the joint angles $\mathbf{\theta}$ are recovered with a standard inverse-kinematics solver~\citep{Zakka_Mink_Python_inverse_2026}. The SVD registration averages prediction noise and projects the result onto the closest valid rigid transformation. Empirically, SVD registration is stable: on Dexonomy, the mean palm registration error is below 0.4 mm for 50\% of grasps and 0.99 mm for 99\%. The IK step projects the generated keypoints onto the hand's kinematic manifold, but it does not use object geometry to optimize contact, penetration, or force closure. Empirically, IK projection is both effective and efficient: on Dexonomy, it achieves an average keypoint reconstruction error of only 0.7 mm and takes 0.002 s per grasp with 64 workers during batched inference (see SUPP for details and visualization). This also indicates that KPGrasp successfully learns the hand structure from data.

\vspace{-2mm}
\subsection{Conditional Keypoint Flow Model}
\label{sec:conditional_keypoint_flow_model}
\vspace{-2mm}

\textbf{Object Encoder.} The input point cloud $\mathbf{O}$ is sampled to 1,024 points and processed by MinkUNet, a 3D sparse convolutional backbone based on Minkowski convolutional networks~\citep{choy20194d}. We then apply Farthest Point Sampling (FPS)~\citep{qi2017pointnetplusplus} to obtain $P'=128$ geometry tokens $\mathbf{T}_{\text{obj}} \in \mathbb{R}^{P' \times C}$, where $C=512$ is the hidden feature dimension. These tokens preserve object geometry relevant to grasp generation while keeping attention computation efficient.

\begin{figure}[t]
    \centering
    \includegraphics[width=0.99\linewidth]{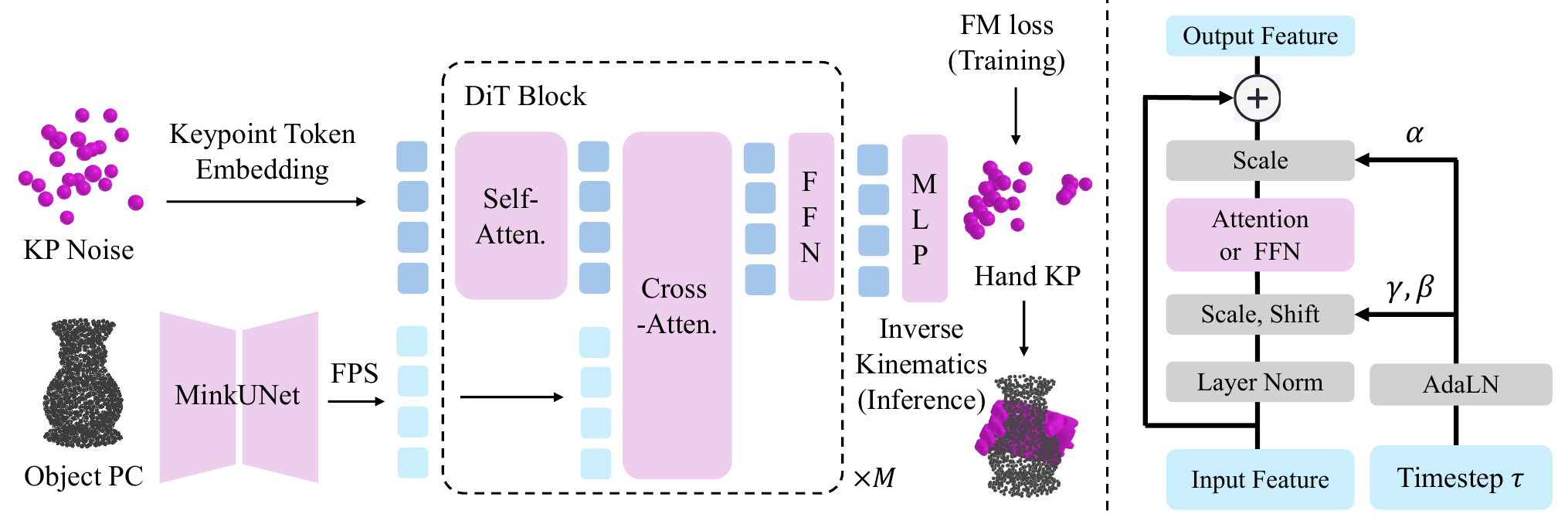}
    \caption{\textbf{KPGrasp Model.} We use a conditional DiT-based network to learn the keypoint flow. The model is trained only with the standard flow-matching objective. During inference, the predicted keypoints are converted into joint angles by inverse kinematics for execution. The right panel shows the adaptive layer normalization that injects the flow time $\tau$ into each attention or FFN sublayer.}
    \label{fig: pipeline}
    \vspace{-3mm}
\end{figure}

\textbf{Keypoint Token Embedding}. Given a noisy keypoint set $\mathbf{K}_\tau \in \mathbb{R}^{J \times 3}$ at time $\tau$, we independently project each keypoint from $\mathbb{R}^3$ to $C$ dimensions with an MLP and add learnable keypoint positional embeddings, resulting in hand keypoint tokens $\mathbf{T}_{\text{kp}} \in \mathbb{R}^{J \times C}$. The flow time $\tau$ is injected separately through adaptive layer normalization in each attention or FFN sublayer, as shown in Figure~\ref{fig: pipeline}.

\textbf{Action Head.} The action head consists of $M=8$ Diffusion Transformer (DiT) blocks~\citep{peebles2023scalable}.  Each DiT block uses 8 attention heads and an FFN hidden dimension of 2,048. Each DiT block alternates between keypoint self-attention and object-conditioned cross-attention. Self-attention operates on $\mathbf{T}_{\text{kp}}$ to model correlations among hand keypoints, while cross-attention uses $\mathbf{T}_{\text{kp}}$ as queries and $\mathbf{T}_{\text{obj}}$ as keys and values to align the hand geometry with the object surface. A final MLP maps the updated $\mathbf{T}_{\text{kp}}$ tokens to the velocity field $\mathbf{v} \in \mathbb{R}^{J \times 3}$ used by flow matching. 

\textbf{Training.} KPGrasp is trained using only the standard flow-matching objective in Eq.~\ref{eq:loss} and the model contains 67.19M parameters. We use the Muon optimizer~\citep{liu2025muon} with a learning rate of $2 \times 10^{-4}$ and train for 200{,}000 steps with a global batch size of 4{,}096 across 8 NVIDIA RTX 4090 GPUs. Training takes approximately two days.

\textbf{Inference.} We draw a batch of $N=100$ Gaussian keypoint-noise samples and solve the learned ODE with 5 Euler integration steps. For each generated candidate, we estimate its likelihood using the Hutchinson trace estimator with one Rademacher probe vector per ODE trajectory and select the 10 highest-likelihood grasps. The score is not a physical stability metric; instead, it favors samples that are likely under the learned grasp distribution and is computed from the same flow model without additional training data or auxiliary networks.
Finally, the generated keypoints are mapped to $[\mathbf{R}, \mathbf{t}, \boldsymbol{\theta}]$ using the rigid-registration and IK procedure described in Sec.~\ref{sec:keypoint_grasp_parameterization}, and then used for simulation or real-robot execution.

\vspace{-3mm}
\section{Experiment}
\vspace{-3mm}

We evaluate KPGrasp on two simulation benchmarks, Dexonomy~\citep{chen2025dexonomy} and DexGrasp Anything~\citep{zhong2025dexgrasp}, and further validate it in real-robot experiments. In simulation, we use complete object point clouds in the object frame to isolate grasp generation and match offline grasp synthesis, where analytic optimization remains dominant. KPGrasp can thus be viewed as an amortized grasp-synthesis model from complete geometry. In real-world experiments, we retrain on simulated single-view point clouds to verify that the same architecture also works with partial observations. Unless otherwise specified, each baseline follows its official inference and selection protocol. Methods with built-in ranking (e.g., Dexonomy) retain that component, while methods (e.g., DexGrasp Anything) without such selection are not given an additional ranker, which would change their native inference pipeline.

\vspace{-2mm}
\subsection{Benchmark on Dexonomy}
\vspace{-2mm}

\textbf{Experimental Settings.} The official Dexonomy dataset contains 10.7k objects and 9.5M grasps covering 31 grasp types. Its diverse grasp taxonomy induces a highly multi-modal grasp distribution, making it a challenging benchmark for generative models. We follow the official object-level train/test split. All baselines are retrained on the official Dexonomy dataset, except the classifier-based Dexonomy variant. While both Dexonomy variants (w/ and w/o classifier) rank 100 candidates by flow likelihood and retain 10 as the final prediction, the classifier variant additionally predicts suitable grasp types and introduces a second selection stage, which is therefore not directly comparable.

\textbf{Evaluation Metrics.} We report the official Dexonomy metrics: grasp success rate (GSR), object success rate (OSR), contact distance consistency (CDC), penetration depth (PD), and diversity. GSR measures the fraction of generated grasps that pass the force test, while OSR measures the fraction of objects with at least one successful grasp. A generated grasp is considered successful only if the object remains stable under all six orthogonal external forces in MuJoCo~\citep{todorov2012mujoco}. CDC, PD, and diversity also follow Dexonomy; lower values indicate more consistent contact distance among fingers, less hand-object penetration, and less collapse along the first principal component, respectively.

\textbf{Quantitative Results.} As shown in Table~\ref{tab: baseline}, KPGrasp achieves a GSR of 76.3\% and an OSR of 97.3\%, improving over the strongest directly comparable baseline by 47.4 percentage points in GSR. KPGrasp even outperforms the classifier-based Dexonomy variant by 12.4\% in GSR without additional generated training data or a separate classifier. It also reduces penetration depth to 2.4 mm and improves CDC, indicating that the generated grasps are more stable and geometrically cleaner.

\begin{table}[t]
\centering
\caption{\textbf{Benchmark on Dexonomy.} KPGrasp outperforms baselines on all metrics. The gray row uses a classifier, which needs extra training data and is thus not directly comparable.}
\label{tab: baseline}
\resizebox{0.9\linewidth}{!}{
\begin{tabular}{lccccc}
\toprule
Method  & GSR($\%$)$\uparrow$ & OSR($\%$)$\uparrow$ & CDC($mm$)$\downarrow$ & PD($mm$)$\downarrow$ & Diversity$\downarrow$ \\
\midrule
UniDexGrasp~\citep{xu2023unidexgrasp} & 24.5
&73.2
&15.6
&11.6
&28.0
\\
DexGrasp Anything~\citep{zhong2025dexgrasp} &21.8
&72.2
&11.0
&14.2
&23.3\\
Dexonomy (w/o classifier)~\citep{chen2025dexonomy} &28.9
&78.1
&14.8
&11.2
&24.6\\
\color{gray}Dexonomy (w/ classifier)&\color{gray}63.9
&\color{gray}91.3
&\color{gray}13.9
&\color{gray}8.6
&\color{gray}25.7\\
\midrule
KPGrasp (Ours) &\textbf{76.3}
&\textbf{97.3}
&\textbf{5.6}
&\textbf{2.4}
&\textbf{23.3}\\
\bottomrule
\end{tabular}
}
\vspace{-3mm}
\end{table}

\begin{figure}
    \centering
    \includegraphics[width=\linewidth]{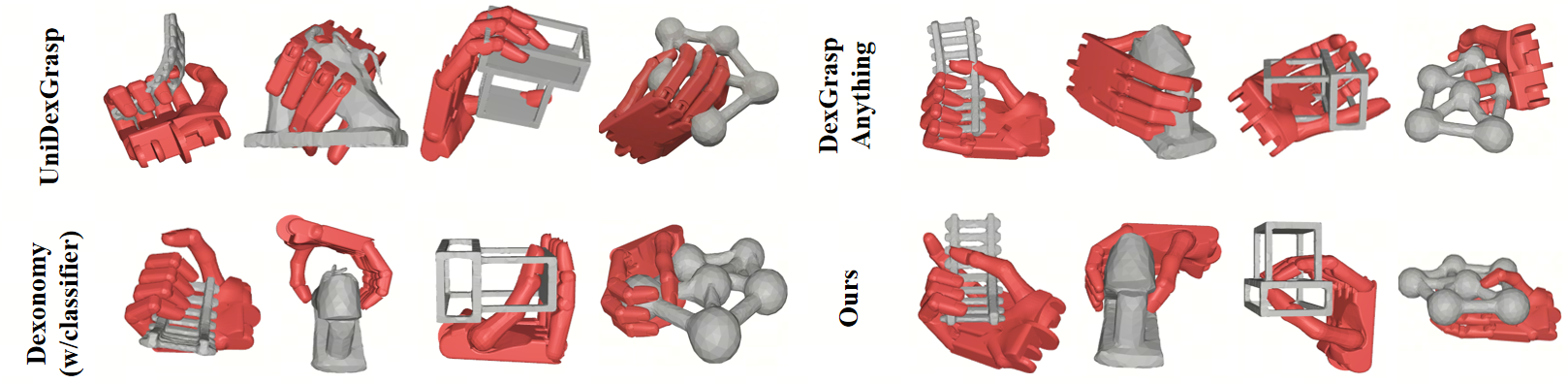}
    \caption{\textbf{Qualitative Comparison on Dexonomy.} KPGrasp produces stable and diverse grasps on complex objects, while baselines often have severe hand-object penetration.}
    \label{fig:comparison}
    \vspace{-3mm}
\end{figure}

\textbf{Qualitative Results.} As shown in Fig.~\ref{fig:comparison}, UniDexGrasp and DexGrasp Anything often produce severe penetration on complex object geometries. Although DexGrasp Anything uses contact and penetration guidance during both training and inference, such geometric objectives can be difficult to tune for complex shapes and may interfere with generative model convergence. In contrast, KPGrasp learns a general grasp prior from large-scale, high-quality data, enabling it to directly generate grasps with accurate contact and low penetration without contact-based test-time refinement. %Our model also captures local geometries and generates similar grasps for similar local geometries as shown in Fig.~\ref{fig: gallery}.

\begin{table*}[b]
\vspace{-4mm}
\centering
\caption{\textbf{Benchmark on DexGrasp Anything.} GRAB, RealD, MD, UDG, and DGN denote DexGRAB~\citep{zhong2025dexgrasp}, RealDex~\citep{liu2024realdex}, MultiDex~\citep{li2022gendexgrasp}, UniDexGrasp~\citep{xu2023unidexgrasp}, and DexGraspNet~\citep{wang2022dexgraspnet}. \textbf{Bold} indicates the best performance, and \underline{underline} indicates the second best. KPGrasp is trained on Dexonomy and evaluated without fine-tuning or contact-based test-time refinement.}
\label{tab: baseline2}
\resizebox{\linewidth}{!}{
\begin{tabular}{lcccccccccccc}
\toprule
\multirow{2}{*}{Method} & \multicolumn{6}{c}{Penetration Depth (mm)} & \multicolumn{6}{c}{Success Rate (\%)} \\
\cmidrule(lr){2-7} \cmidrule(lr){8-13}
 & GRAB & RealD & MD & UDG & DGN & Avg. & GRAB & RealD & MD & UDG & DGN & Avg. \\
\midrule
UniDexGrasp~\citep{xu2023unidexgrasp} & 37.4 & 39.0 & 13.5 & 24.5 & 31.9 & 29.3 & 20.8 & 27.1 & 21.6 & 23.7 & 33.9 & 25.4 \\
GraspTTA~\citep{jiang2021hand} & 51.4 & 40.1 & 19.0 & 21.2 & 24.5 & 31.2 & 14.4 & 13.3 & 30.3 & 21.0 & 18.6 & 19.5 \\
SceneDiffuser~\citep{huang2023diffusion} & 41.1 & 42.0 & 14.6 & 25.1 & 31.0 & 30.8 & 39.1 & 21.7 & \underline{69.8} & 28.3 & 26.6 & 37.1 \\
UGG~\citep{lu2024ugg} & 33.2 & 34.4 & \underline{10.3} & 24.5 & 25.2 & 25.5 & 42.7 & 32.7 & 55.3 & 46.0 & 46.9 & 44.7 \\
DexGraspAnything~\citep{zhong2025dexgrasp} & \underline{30.4} & \underline{27.7} & 11.4 & \underline{18.8} & \underline{17.8} & \underline{21.2} & \underline{57.9} & \underline{44.8} & \textbf{79.1} & \underline{53.1} & \underline{57.5} & \underline{58.5} \\
\textbf{KPGrasp (Ours)} & \textbf{20.8} & \textbf{9.6} & \textbf{6.6} & \textbf{13.0} & \textbf{14.3} & \textbf{12.9} & \textbf{59.8} & \textbf{67.1} & 62.2 & \textbf{62.3} & \textbf{67.5} & \textbf{63.8} \\
\bottomrule
\end{tabular}
}
\vspace{-3mm}
\end{table*}

\vspace{-2mm}
\subsection{Benchmark on DexGrasp Anything}
\vspace{-2mm}

\textbf{Experimental Settings.} We next evaluate cross-benchmark generalization on the DexGrasp Anything benchmark~\citep{zhong2025dexgrasp}. KPGrasp is trained on Dexonomy and tested on the official DexGrasp Anything object sets without fine-tuning or contact-based test-time refinement. We ensure that the Dexonomy training objects do not overlap with the evaluation objects. For the results of training KPGrasp on the DexGrasp Anything benchmark, we report and analyze them in SUPP.

\textbf{Evaluation Metrics.} We follow the DexGrasp Anything evaluation protocol and use its official code to test our model. Baseline results are quoted from the DexGrasp Anything paper under the same published protocol. For conciseness, we report the two most important metrics: Suc.6 as the success rate and penetration depth in millimeters. Note that this protocol differs from Dexonomy: success is evaluated in IsaacGym~\citep{makoviychuk2021isaac}, and penetration is computed from point clouds and point normals.

\textbf{Quantitative Results.} Table~\ref{tab: baseline2} shows that KPGrasp obtains the best average success rate and the lowest average penetration depth across the five datasets, despite being trained only on Dexonomy. This suggests that the keypoint flow model learns a transferable grasp prior rather than overfitting to a single benchmark protocol. Baselines with task-specific heuristics can win on individual datasets, but their gains do not transfer uniformly across datasets. To assess whether the improvement is explained by the training dataset,
we retrain DGA on Dexonomy and evaluate it under the official DGA protocol.
Its average success rate increases only from 58.5\% to 59.1\%.
Without contact-based test-time refinement, DGA achieves 46.6\% and
46.8\% when trained on its original datasets and Dexonomy, respectively,
compared with 63.8\% for KPGrasp. Thus, changing the training dataset alone does not account for
KPGrasp's advantage. % The penetration reported here is worse than on Dexonomy partly because the point-normal-based penetration calculation in DexGrasp Anything can be sensitive to abnormal object normals for some meshes (see SUPP for details).

\begin{table}[t]
    \caption{\textbf{Ablation Study on Parameterization, Tokenization, and Likelihood Ranking.}}
    \centering
    \resizebox{0.9\linewidth}{!}{
    \begin{tabular}{lccccc}
    \toprule
    Method & GSR($\%$)$\uparrow$ & OSR($\%$)$\uparrow$& CDC($mm$)$\downarrow$& PD($mm$)$\downarrow$ & Diversity$\downarrow$ \\
    \midrule
        Pose-Joint (w/ Euclidean Flow) & 62.5 & 92.0 & 10.3 & 3.7 & 25.7  \\
        Pose-Joint (w/ $SE(3)$ Flow) & 55.0 & 93.9 & 11.8 & 5.7 & 23.8\\
        \midrule
        Pose-Joint (w/ 24 tokens) & 63.5 & 94.6 & 9.7 & 3.3 & 25.8 \\
        Pose-Joint (w/ 34 tokens) & 66.1 & 95.1 & 9.5 & 3.2 & 25.7 \\
        Keypoint (w/ Single Token) & 55.4 & 93.2 & 10.1 & 8.6 & 23.7 \\
        \midrule
        Keypoint (w/o likelihood ranking) & 67.0 & 96.4 & 7.4 & 4.4 & \textbf{23.2} \\
        Keypoint (Ours) & \textbf{76.3} & \textbf{97.3} & \textbf{5.6} & \textbf{2.4} & 23.3 \\
    \bottomrule
    \end{tabular}
    }
    \label{tab: param ablation}
    \vspace{-4mm}
\end{table}

\begin{figure}
    \centering
    \includegraphics[width=0.99\linewidth]{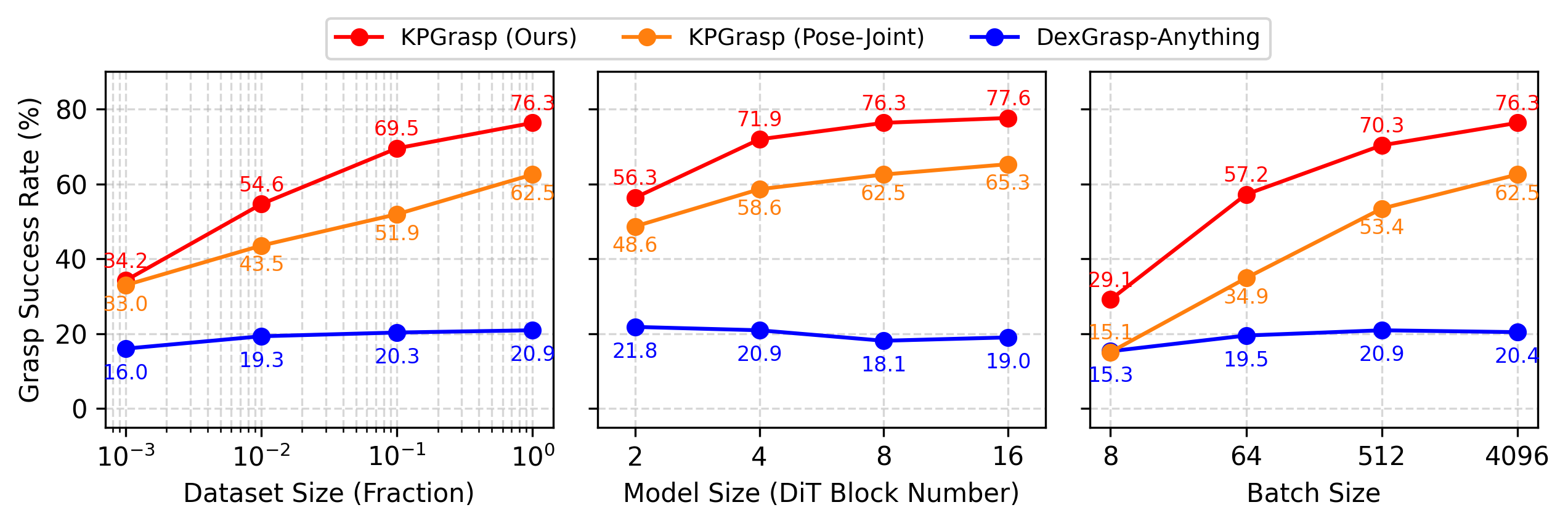}
    \caption{\textbf{Scaling Behavior.} KPGrasp improves as training data, model size, and batch size increase. Keypoint parameterization consistently outperforms the conventional pose-joint variant.}
    \label{fig: scaling}
    \vspace{-3mm}
\end{figure}

\vspace{-3mm}
\subsection{Ablation Study}
\vspace{-2mm}

The ablation studies are performed on the Dexonomy benchmark to isolate the effect of hand parameterization, tokenization, likelihood ranking, and scaling. 

\textbf{Hand Parameterization.} We replace the keypoint output space with the pose-joint parameterization with other settings unchanged. The first variant represents wrist rotation as a 9D Euclidean matrix during flow matching and projects it to $SO(3)$ with SVD~\citep{levinson2020analysis} at the end of inference. The second variant applies $SE(3)$ flow matching~\citep{bose2023se} to the wrist pose. 
As shown in Table~\ref{tab: param ablation}, the keypoint parameterization outperforms both pose-joint variants. The $SE(3)$ flow variant performs worse than the simpler 9D Euclidean rotation variant, which may reflect the additional tuning difficulty of manifold flow matching and is consistent with observations in prior work~\citep{chisari2024learning}.  

\textbf{Tokenization.} We further ablate tokenization to separate the effect of hand parameterization from the number of output tokens. For pose-joint outputs, the 24-token variant uses one token each for the 9D rotation and 3D translation, plus 22 joint-angle tokens; the 34-token variant uses one token per scalar (3+9+22); for keypoints, we collapse all keypoints into a single token. As shown in Table~\ref{tab: param ablation}, pose-joint outputs remain substantially worse than our model even with multiple tokens, while keypoint outputs degrade sharply when represented by a single token. This suggests that tokenization matters but token count alone cannot explain the gain, and that KPGrasp, with its Euclidean keypoint representation, greatly benefits from reasoning among multiple spatial keypoint tokens.

\textbf{Likelihood Ranking.} Next, we compare against randomly sampling 10 grasps out of 100 candidates without the likelihood ranking described in Sec.~\ref{sec:conditional_keypoint_flow_model}. Since the generated candidates are iid before ranking, this setting is distributionally equivalent to generating 10 grasps directly without any selection. As shown in Table~\ref{tab: param ablation}, likelihood ranking improves GSR from 67.0\% to 76.3\%, verifying that the flow likelihood provides an effective score for selecting generated candidates. 

\textbf{Scaling Behavior.} We vary dataset size, model size, and batch size separately while keeping the remaining settings fixed. Dataset fractions are $[0.001, 0.01, 0.1, 1]$, where the smallest subset has roughly 10k randomly sampled grasps. Model size is controlled by the number of DiT blocks $M \in [2,4,8,16]$. Batch size ranges from $[8, 64, 512, 4096]$, and the total number of training steps is fixed. With the total number of training samples held fixed, training with a batch size of 512 requires 1.6M steps (approximately 13 days) and yields a success rate of $73.5\%$, further supporting our conclusion. Figure~\ref{fig: scaling} shows two key observations. First, KPGrasp scales effectively: increasing data, model, and batch size improves success rate, whereas the DexGrasp Anything baseline does not. Second, the keypoint parameterization consistently outperforms the pose-joint variant with Euclidean flow across all scaling settings, with the gap widening as scale increases. These results suggest that both the KPGrasp architecture and the keypoint parameterization are important for the final performance.

\vspace{-2mm}
\subsection{Real World Experiments}
\vspace{-2mm}

\begin{wrapfigure}{R}{0.4\textwidth}
    \vspace{-5mm}
    \centering
    \includegraphics[width=\linewidth]{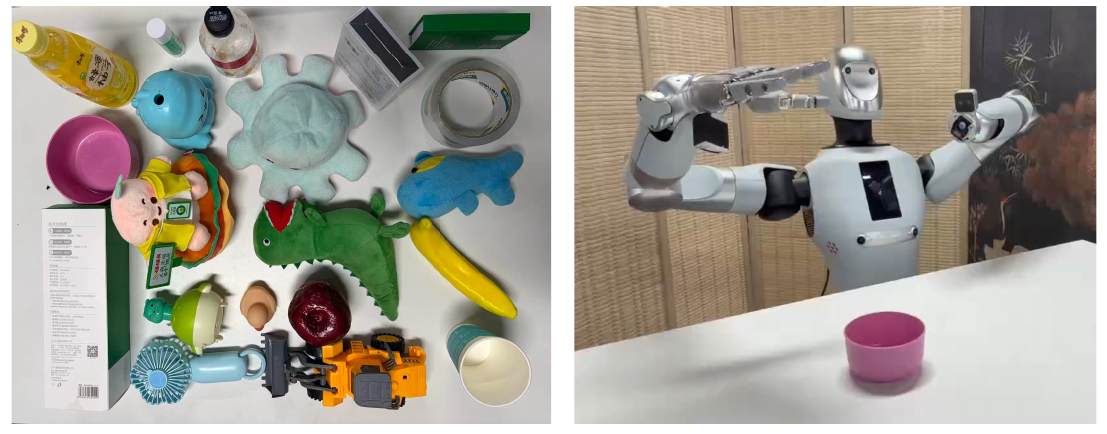}
    \caption{\textbf{Real World Setup.} (Left) 20 test objects. (Right) Robot setup.}
    \label{fig: real-world}
    \vspace{-4mm}
\end{wrapfigure}

Finally, we evaluate KPGrasp on partial point clouds in the real world. Figure~\ref{fig: real-world} shows the 20 test objects with diverse shapes and sizes, and the robot setup of a Galbot platform with a Sharpa right hand and a RealSense D405 camera mounted on the left arm.

Following the Dexonomy protocol, we regenerate tabletop grasp data with its analytic pipeline and retrain KPGrasp for this setup. The training set contains 3.3M grasps over 5k DGN objects with simulated single-view point clouds. At test time, we segment a single-view point cloud in the robot base frame using GroundedSAM~\citep{ren2024grounded}, generate 100 grasp candidates, rank them by likelihood, filter collisions, and execute the best remaining grasp by trajectory interpolation. We run 5 trials per object for 100 trials in total, achieving 83/100 successes. A trial is successful if the object remains held after lifting. A major failure mode is that some generated grasps are sensitive to small object-pose perturbations, which can arise from calibration errors and sensor noise. Please see the supplementary video for grasp trials.

\vspace{-2mm}
\subsection{Inference Time Analysis}
\vspace{-2mm}

We measure end-to-end inference, including network sampling, likelihood ranking, and inverse kinematics, on an NVIDIA RTX 4090 GPU. KPGrasp samples 100 candidates per object, returns the top 10 by likelihood, and processes 4 objects per GPU batch, taking 0.032 s per returned grasp. This is around 87 times faster than DexGrasp Anything, which uses contact-based test-time refinement and takes around 2.8 s per grasp.

KPGrasp also provides higher valid-grasp throughput than analytic data-generation pipelines. Dexonomy~\citep{chen2025dexonomy} reports generating 9.5M grasps in more than 2 days on 8 GPUs, corresponding to less than 7 valid grasps per second per GPU. Since KPGrasp produces one grasp every 0.032 s and achieves a 76.3\% success rate on Dexonomy, its effective throughput is approximately $(1/0.032)\times0.763\approx24$ valid grasps per second, more than 3 times faster than Dexonomy-style analytic generation.

\vspace{-3mm}
\section{Limitation}
\vspace{-3mm}

First, because KPGrasp learns grasp priors implicitly from data, its performance depends strongly on the quality and coverage of the training dataset. It may therefore degrade when the available data are noisy, biased, or limited in scale, which is common in many other robotic tasks like manipulation. Second, our real-world experiments are intended to validate deployability under partial observations, and they are still not a comprehensive real-world benchmark. Finally, learning a unified model across different robotic hands remains an open direction.

\vspace{-3mm}
\section{Conclusion}
\vspace{-3mm}

We presented KPGrasp, a flow-matching framework for dexterous grasp generation from object point clouds. By combining the keypoint parameterization and a simple yet scalable Transformer-based conditional flow model, KPGrasp directly learns the grasp prior from large-scale data without explicit contact losses or contact-based test-time refinement. Experiments show strong simulation performance, efficient inference, favorable scaling behavior, and successful real-world deployment.

\begin{figure}[h]
    \centering
    \includegraphics[width=0.99\linewidth]{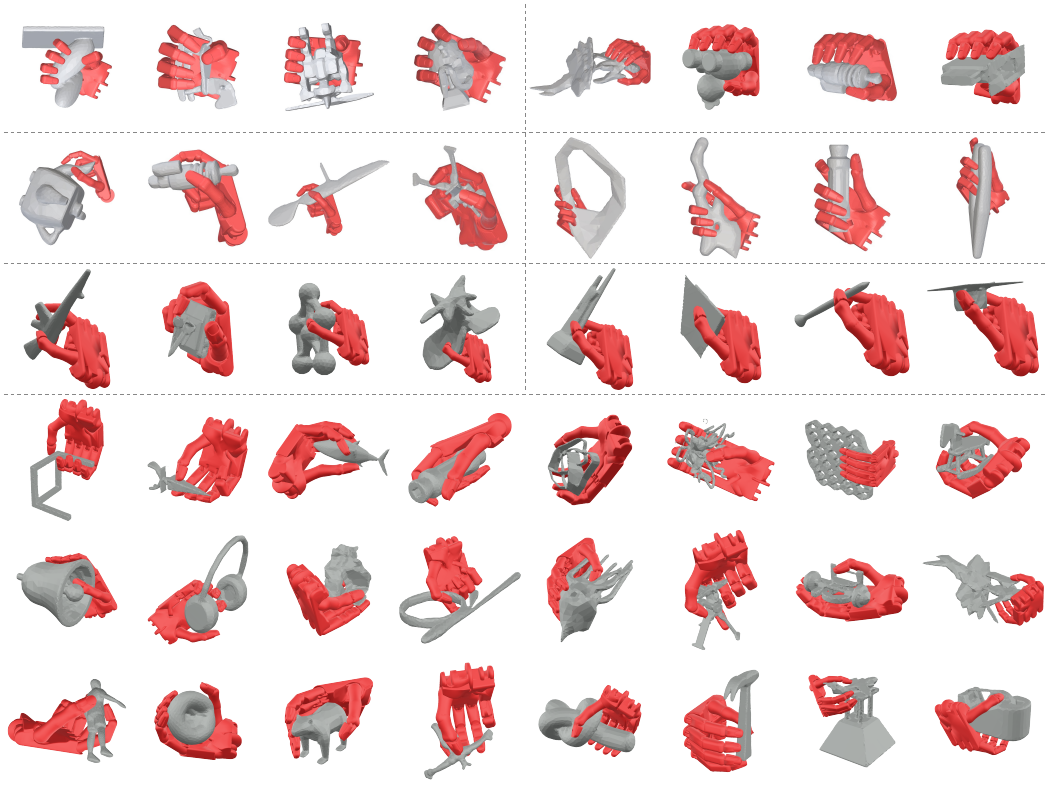}
    \caption{\textbf{Grasp Gallery.} KPGrasp generates similar grasps for objects with the same local geometries. This also shows more examples of high-quality grasps for complex geometries.}
    \label{fig: gallery}
    \vspace{-3mm}
\end{figure}

\clearpage
% The acknowledgments are automatically included only in the final and preprint versions of the paper.
%\acknowledgments{We thank the anonymous reviewers and area chair for their constructive feedback.}

% The CoRL style file selects corlabbrvnat automatically.
\bibliography{main}

@String(AAAI  = {AAAI})

@article{feix2015grasp,
  title={The grasp taxonomy of human grasp types},
  author={Feix, Thomas and Romero, Javier and Schmiedmayer, Heinz-Bodo and Dollar, Aaron M and Kragic, Danica},
  journal={IEEE Transactions on human-machine systems},
  volume={46},
  number={1},
  pages={66--77},
  year={2015},
  publisher={IEEE}
}

@inproceedings{ferrari1992planning,
  title={Planning optimal grasps},
  author={Ferrari, Carlo and Canny, John and others},
  booktitle={Proceedings., 1992 IEEE International Conference on Robotics and Automation, 1992.},
  volume={3},
  pages={2290--2295},
  year={1992},
  organization={IEEE}
}

@article{miller2004graspit,
  title={Graspit! a versatile simulator for robotic grasping},
  author={Miller, Andrew T and Allen, Peter K},
  journal={IEEE Robotics \& Automation Magazine},
  volume={11},
  number={4},
  pages={110--122},
  year={2004},
  publisher={IEEE}
}

@inproceedings{ciocarlie2007dexterous,
  title={Dexterous grasping via eigengrasps: A low-dimensional approach to a high-complexity problem},
  author={Ciocarlie, Matei and Goldfeder, Corey and Allen, Peter},
  booktitle={Robotics: Science and systems manipulation workshop-sensing and adapting to the real world},
  year={2007}
}

@article{liu2021synthesizing,
  title={Synthesizing diverse and physically stable grasps with arbitrary hand structures using differentiable force closure estimator},
  author={Liu, Tengyu and Liu, Zeyu and Jiao, Ziyuan and Zhu, Yixin and Zhu, Song-Chun},
  journal={IEEE Robotics and Automation Letters},
  volume={7},
  number={1},
  pages={470--477},
  year={2021},
  publisher={IEEE}
}

@article{wang2022dexgraspnet,
  title={Dexgraspnet: A large-scale robotic dexterous grasp dataset for general objects based on simulation},
  author={Wang, Ruicheng and Zhang, Jialiang and Chen, Jiayi and Xu, Yinzhen and Li, Puhao and Liu, Tengyu and Wang, He},
  journal={arXiv preprint arXiv:2210.02697},
  year={2022}
}

@inproceedings{li2023frogger,
  title={Frogger: Fast robust grasp generation via the min-weight metric},
  author={Li, Albert H and Culbertson, Preston and Burdick, Joel W and Ames, Aaron D},
  booktitle={2023 IEEE/RSJ International Conference on Intelligent Robots and Systems (IROS)},
  pages={6809--6816},
  year={2023},
  organization={IEEE}
}

@article{chen2024springgrasp,
  title={SpringGrasp: An optimization pipeline for robust and compliant dexterous pre-grasp synthesis},
  author={Chen, Sirui and Bohg, Jeannette and Liu, C Karen},
  journal={arXiv preprint arXiv:2404.13532},
  year={2024}
}

@article{liu2024realdex,
  title={Realdex: Towards human-like grasping for robotic dexterous hand},
  author={Liu, Yumeng and Yang, Yaxun and Wang, Youzhuo and Wu, Xiaofei and Wang, Jiamin and Yao, Yichen and Schwertfeger, S{\"o}ren and Yang, Sibei and Wang, Wenping and Yu, Jingyi and others},
  journal={arXiv preprint arXiv:2402.13853},
  year={2024}
}

@inproceedings{todorov2012mujoco,
  title={MuJoCo: A physics engine for model-based control},
  author={Todorov, Emanuel and Erez, Tom and Tassa, Yuval},
  booktitle={2012 IEEE/RSJ International Conference on Intelligent Robots and Systems},
  pages={5026--5033},
  year={2012},
  organization={IEEE},
  doi={10.1109/IROS.2012.6386109}
}

@article{makoviychuk2021isaac,
  title={Isaac gym: High performance gpu-based physics simulation for robot learning},
  author={Makoviychuk, Viktor and Wawrzyniak, Lukasz and Guo, Yunrong and Lu, Michelle and Storey, Kier and Macklin, Miles and Hoeller, David and Rudin, Nikita and Allshire, Arthur and Handa, Ankur and others},
  journal={arXiv preprint arXiv:2108.10470},
  year={2021}
}

@software{Zakka_Mink_Python_inverse_2026,
  author = {Zakka, Kevin},
  title = {{Mink: Python inverse kinematics based on MuJoCo}},
  year = {2026},
  month = feb,
  version = {1.1.0},
  url = {https://github.com/kevinzakka/mink},
  license = {Apache-2.0}
}

@inproceedings{chen2024task,
  title={Task-oriented dexterous hand pose synthesis using differentiable grasp wrench boundary estimator},
  author={Chen, Jiayi and Chen, Yuxing and Zhang, Jialiang and Wang, He},
  booktitle={2024 IEEE/RSJ International Conference on Intelligent Robots and Systems (IROS)},
  pages={5281--5288},
  year={2024},
  organization={IEEE}
}

@article{chen2025dexonomy,
  title={Dexonomy: Synthesizing all dexterous grasp types in a grasp taxonomy},
  author={Chen, Jiayi and Ke, Yubin and Peng, Lin and Wang, He},
  journal={arXiv preprint arXiv:2504.18829},
  year={2025}
}

@inproceedings{chen2025bodex,
  title={Bodex: Scalable and efficient robotic dexterous grasp synthesis using bilevel optimization},
  author={Chen, Jiayi and Ke, Yubin and Wang, He},
  booktitle={2025 IEEE International Conference on Robotics and Automation (ICRA)},
  pages={01--08},
  year={2025},
  organization={IEEE}
}

@inproceedings{he2025dexvlg,
  title={Dexvlg: Dexterous vision-language-grasp model at scale},
  author={He, Jiawei and Li, Danshi and Yu, Xinqiang and Qi, Zekun and Zhang, Wenyao and Chen, Jiayi and Zhang, Zhaoxiang and Zhang, Zhizheng and Yi, Li and Wang, He},
  booktitle={Proceedings of the IEEE/CVF International Conference on Computer Vision},
  pages={14248--14258},
  year={2025}
}

@inproceedings{zhang2024dexgraspnet,
  title={Dexgraspnet 2.0: Learning generative dexterous grasping in large-scale synthetic cluttered scenes},
  author={Zhang, Jialiang and Liu, Haoran and Li, Danshi and Yu, XinQiang and Geng, Haoran and Ding, Yufei and Chen, Jiayi and Wang, He},
  booktitle={8th Annual Conference on Robot Learning},
  year={2024}
}

@inproceedings{zhou2019continuity,
  title={On the continuity of rotation representations in neural networks},
  author={Zhou, Yi and Barnes, Connelly and Lu, Jingwan and Yang, Jimei and Li, Hao},
  booktitle={Proceedings of the IEEE/CVF conference on computer vision and pattern recognition},
  pages={5745--5753},
  year={2019}
}

@article{umeyama1991least,
  title={Least-squares estimation of transformation parameters between two point patterns},
  author={Umeyama, Shinji},
  journal={IEEE Transactions on Pattern Analysis and Machine Intelligence},
  volume={13},
  number={4},
  pages={376--380},
  year={1991},
  publisher={IEEE}
}

@article{liu2020deep,
  title={Deep differentiable grasp planner for high-dof grippers},
  author={Liu, Min and Pan, Zherong and Xu, Kai and Ganguly, Kanishka and Manocha, Dinesh},
  journal={arXiv preprint arXiv:2002.01530},
  year={2020}
}

@article{chen2022learning,
  title={Learning robust real-world dexterous grasping policies via implicit shape augmentation},
  author={Chen, Zoey Qiuyu and Van Wyk, Karl and Chao, Yu-Wei and Yang, Wei and Mousavian, Arsalan and Gupta, Abhishek and Fox, Dieter},
  journal={arXiv preprint arXiv:2210.13638},
  year={2022}
}

@inproceedings{xu2023unidexgrasp,
  title={Unidexgrasp: Universal robotic dexterous grasping via learning diverse proposal generation and goal-conditioned policy},
  author={Xu, Yinzhen and Wan, Weikang and Zhang, Jialiang and Liu, Haoran and Shan, Zikang and Shen, Hao and Wang, Ruicheng and Geng, Haoran and Weng, Yijia and Chen, Jiayi and others},
  booktitle={Proceedings of the IEEE/CVF Conference on Computer Vision and Pattern Recognition},
  pages={4737--4746},
  year={2023}
}

@inproceedings{xu2024dexterous,
  title={Dexterous grasp transformer},
  author={Xu, Guo-Hao and Wei, Yi-Lin and Zheng, Dian and Wu, Xiao-Ming and Zheng, Wei-Shi},
  booktitle={Proceedings of the IEEE/CVF Conference on Computer Vision and Pattern Recognition},
  pages={17933--17942},
  year={2024}
}

@inproceedings{lu2024ugg,
  title={Ugg: Unified generative grasping},
  author={Lu, Jiaxin and Kang, Hao and Li, Haoxiang and Liu, Bo and Yang, Yiding and Huang, Qixing and Hua, Gang},
  booktitle={European Conference on Computer Vision},
  pages={414--433},
  year={2024},
  organization={Springer}
}

@article{wei2024mathcal,
  title={D (R, O) Grasp: A Unified Representation of Robot and Object Interaction for Cross-Embodiment Dexterous Grasping},
  author={Wei, Zhenyu and Xu, Zhixuan and Guo, Jingxiang and Hou, Yiwen and Gao, Chongkai and Cai, Zhehao and Luo, Jiayu and Shao, Lin},
  journal={arXiv preprint arXiv:2410.01702},
  year={2024}
}

@article{feng2024ffhflow,
  title={FFHFlow: Diverse and Uncertainty-Aware Dexterous Grasp Generation via Flow Variational Inference},
  author={Feng, Qian and Feng, Jianxiang and Chen, Zhaopeng and Triebel, Rudolph and Knoll, Alois},
  journal={arXiv preprint arXiv:2407.15161},
  year={2024}
}

@inproceedings{zhong2025dexgrasp,
  title={Dexgrasp anything: Towards universal robotic dexterous grasping with physics awareness},
  author={Zhong, Yiming and Jiang, Qi and Yu, Jingyi and Ma, Yuexin},
  booktitle={Proceedings of the Computer Vision and Pattern Recognition Conference},
  pages={22584--22594},
  year={2025}
}

@inproceedings{huang2023diffusion,
  title={Diffusion-based generation, optimization, and planning in 3d scenes},
  author={Huang, Siyuan and Wang, Zan and Li, Puhao and Jia, Baoxiong and Liu, Tengyu and Zhu, Yixin and Liang, Wei and Zhu, Song-Chun},
  booktitle={Proceedings of the IEEE/CVF Conference on Computer Vision and Pattern Recognition},
  pages={16750--16761},
  year={2023}
}

@inproceedings{zhu2021toward,
  title={Toward human-like grasp: Dexterous grasping via semantic representation of object-hand},
  author={Zhu, Tianqiang and Wu, Rina and Lin, Xiangbo and Sun, Yi},
  booktitle={Proceedings of the IEEE/CVF International Conference on Computer Vision},
  pages={15741--15751},
  year={2021}
}

@inproceedings{jiang2021hand,
  title={Hand-object contact consistency reasoning for human grasps generation},
  author={Jiang, Hanwen and Liu, Shaowei and Wang, Jiashun and Wang, Xiaolong},
  booktitle={Proceedings of the IEEE/CVF international conference on computer vision},
  pages={11107--11116},
  year={2021}
}

@article{ye2025dex1b,
  title={Dex1b: Learning with 1b demonstrations for dexterous manipulation},
  author={Ye, Jianglong and Wang, Keyi and Yuan, Chengjing and Yang, Ruihan and Li, Yiquan and Zhu, Jiyue and Qin, Yuzhe and Zou, Xueyan and Wang, Xiaolong},
  journal={arXiv preprint arXiv:2506.17198},
  year={2025}
}

@article{li2022gendexgrasp,
  title={Gendexgrasp: Generalizable dexterous grasping},
  author={Li, Puhao and Liu, Tengyu and Li, Yuyang and Geng, Yiran and Zhu, Yixin and Yang, Yaodong and Huang, Siyuan},
  journal={arXiv preprint arXiv:2210.00722},
  year={2022}
}

@article{weng2024dexdiffuser,
  title={Dexdiffuser: Generating dexterous grasps with diffusion models},
  author={Weng, Zehang and Lu, Haofei and Kragic, Danica and Lundell, Jens},
  journal={IEEE Robotics and Automation Letters},
  volume={9},
  number={12},
  pages={11834--11840},
  year={2024},
  publisher={IEEE}
}

@inproceedings{cheng2021handfoldingnet,
  title={Handfoldingnet: A 3d hand pose estimation network using multiscale-feature guided folding of a 2d hand skeleton},
  author={Cheng, Wencan and Park, Jae Hyun and Ko, Jong Hwan},
  booktitle={Proceedings of the IEEE/CVF international conference on computer vision},
  pages={11260--11269},
  year={2021}
}

@inproceedings{chen2023tracking,
  title={Tracking and reconstructing hand object interactions from point cloud sequences in the wild},
  author={Chen, Jiayi and Yan, Mi and Zhang, Jiazhao and Xu, Yinzhen and Li, Xiaolong and Weng, Yijia and Yi, Li and Song, Shuran and Wang, He},
  booktitle={Proceedings of the AAAI conference on artificial intelligence},
  volume={37},
  number={1},
  pages={304--312},
  year={2023}
}

@inproceedings{cheng2024handdiff,
  title={Handdiff: 3d hand pose estimation with diffusion on image-point cloud},
  author={Cheng, Wencan and Tang, Hao and Van Gool, Luc and Ko, Jong Hwan},
  booktitle={Proceedings of the IEEE/CVF Conference on Computer Vision and Pattern Recognition},
  pages={2274--2284},
  year={2024}
}

@article{haldar2025point,
  title={Point policy: Unifying observations and actions with key points for robot manipulation},
  author={Haldar, Siddhant and Pinto, Lerrel},
  journal={arXiv preprint arXiv:2502.20391},
  year={2025}
}

@article{goswami2025world,
  title={World Models Can Leverage Human Videos for Dexterous Manipulation},
  author={Goswami, Raktim Gautam and Bar, Amir and Fan, David and Yang, Tsung-Yen and Zhou, Gaoyue and Krishnamurthy, Prashanth and Rabbat, Michael and Khorrami, Farshad and LeCun, Yann},
  journal={arXiv preprint arXiv:2512.13644},
  year={2025}
}

@article{guzey2025dexterity,
  title={Dexterity from Smart Lenses: Multi-Fingered Robot Manipulation with In-the-Wild Human Demonstrations},
  author={Guzey, Irmak and Qi, Haozhi and Urain, Julen and Wang, Changhao and Yin, Jessica and Bodduluri, Krishna and Lambeta, Mike and Pinto, Lerrel and Rai, Akshara and Malik, Jitendra and others},
  journal={arXiv preprint arXiv:2511.16661},
  year={2025}
}

@inproceedings{lim2024equigraspflow,
  title={Equigraspflow: Se (3)-equivariant 6-dof grasp pose generative flows},
  author={Lim, Byeongdo and Kim, Jongmin and Kim, Jihwan and Lee, Yonghyeon and Park, Frank C},
  booktitle={8th Annual Conference on Robot Learning},
  year={2024}
}

@article{zhu2025se,
  title={Se (3)-equivariant diffusion policy in spherical fourier space},
  author={Zhu, Xupeng and Wang, Fan and Walters, Robin and Shi, Jane},
  journal={arXiv preprint arXiv:2507.01723},
  year={2025}
}

@article{shi2025hograspflow,
  title={HOGraspFlow: Exploring Vision-based Generative Grasp Synthesis with Hand-Object Priors and Taxonomy Awareness},
  author={Shi, Yitian and Guo, Zicheng and Wolf, Rosa and Welte, Edgar and Rayyes, Rania},
  journal={arXiv preprint arXiv:2509.16871},
  year={2025}
}

@article{grathwohl2018ffjord,
  title={Ffjord: Free-form continuous dynamics for scalable reversible generative models},
  author={Grathwohl, Will and Chen, Ricky TQ and Bettencourt, Jesse and Sutskever, Ilya and Duvenaud, David},
  journal={arXiv preprint arXiv:1810.01367},
  year={2018}
}

@article{song2020score,
  title={Score-based generative modeling through stochastic differential equations},
  author={Song, Yang and Sohl-Dickstein, Jascha and Kingma, Diederik P and Kumar, Abhishek and Ermon, Stefano and Poole, Ben},
  journal={arXiv preprint arXiv:2011.13456},
  year={2020}
}

@article{lipman2022flow,
  title={Flow matching for generative modeling},
  author={Lipman, Yaron and Chen, Ricky TQ and Ben-Hamu, Heli and Nickel, Maximilian and Le, Matt},
  journal={arXiv preprint arXiv:2210.02747},
  year={2022}
}

@article{boffi2025build,
  title={How to build a consistency model: Learning flow maps via self-distillation},
  author={Boffi, Nicholas M and Albergo, Michael S and Vanden-Eijnden, Eric},
  journal={arXiv preprint arXiv:2505.18825},
  year={2025}
}

@inproceedings{choy20194d,
  title={4D spatio-temporal convnets: Minkowski convolutional neural networks},
  author={Choy, Christopher and Gwak, JunYoung and Savarese, Silvio},
  booktitle={Proceedings of the IEEE/CVF Conference on Computer Vision and Pattern Recognition},
  pages={3075--3084},
  year={2019}
}

@inproceedings{qi2017pointnetplusplus,
  title={PointNet++: Deep hierarchical feature learning on point sets in a metric space},
  author={Qi, Charles R and Yi, Li and Su, Hao and Guibas, Leonidas J},
  booktitle={Advances in neural information processing systems},
  volume={30},
  year={2017}
}

@inproceedings{peebles2023scalable,
  title={Scalable diffusion models with transformers},
  author={Peebles, William and Xie, Saining},
  booktitle={Proceedings of the IEEE/CVF international conference on computer vision},
  pages={4195--4205},
  year={2023}
}

@article{levinson2020analysis,
  title={An analysis of svd for deep rotation estimation},
  author={Levinson, Jake and Esteves, Carlos and Chen, Kefan and Snavely, Noah and Kanazawa, Angjoo and Rostamizadeh, Afshin and Makadia, Ameesh},
  journal={Advances in Neural Information Processing Systems},
  volume={33},
  pages={22554--22565},
  year={2020}
}

@article{bose2023se,
  title={Se (3)-stochastic flow matching for protein backbone generation},
  author={Bose, Avishek Joey and Akhound-Sadegh, Tara and Huguet, Guillaume and Fatras, Kilian and Rector-Brooks, Jarrid and Liu, Cheng-Hao and Nica, Andrei Cristian and Korablyov, Maksym and Bronstein, Michael and Tong, Alexander},
  journal={arXiv preprint arXiv:2310.02391},
  year={2023}
}

@article{chisari2024learning,
  title={Learning robotic manipulation policies from point clouds with conditional flow matching},
  author={Chisari, Eugenio and Heppert, Nick and Argus, Max and Welschehold, Tim and Brox, Thomas and Valada, Abhinav},
  journal={arXiv preprint arXiv:2409.07343},
  year={2024}
}

@article{liu2025muon,
  title={Muon is Scalable for LLM Training},
  author={Liu, Jingyuan and Su, Jianlin and Yao, Xingcheng and Jiang, Zhejun and Lai, Guokun and Du, Yulun and Qin, Yidao and Xu, Weixin and Lu, Enzhe and Yan, Junjie and Chen, Yanru and Zheng, Huabin and Liu, Yibo and Liu, Shaowei and Yin, Bohong and He, Weiran and Zhu, Han and Wang, Yuzhi and Wang, Jianzhou and Dong, Mengnan and Zhang, Zheng and Kang, Yongsheng and Zhang, Hao and Xu, Xinran and Zhang, Yutao and Wu, Yuxin and Zhou, Xinyu and Yang, Zhilin},
  journal={arXiv preprint arXiv:2502.16982},
  year={2025}
}

@article{ren2024grounded,
  title={Grounded SAM: Assembling Open-World Models for Diverse Visual Tasks
},
  author={Tianhe Ren and Shilong Liu and Ailing Zeng and Jing Lin and Kunchang Li and He Cao and Jiayu Chen and Xinyu Huang and Yukang Chen and Feng Yan and Zhaoyang Zeng and Hao Zhang and Feng Li and Jie Yang and Hongyang Li and Qing Jiang and Lei Zhang},
  journal={arXiv preprint arXiv:2401.14159},
  year={2024}
}

% ===== Supplementary material merged as the appendix; original wording retained. =====
\clearpage
\appendix

\section{Inverse-Kinematics Module Analysis}

\paragraph{Implementation Details}
For each grasp, we first estimate the wrist pose by rigidly aligning the canonical palm keypoints with the predicted palm keypoints. We compute the optimal rotation matrix with SVD and recover the translation from the aligned centroids. Given this wrist pose, we solve for the remaining hand configuration using position-only keypoint tasks in Mink~\cite{Zakka_Mink_Python_inverse_2026} with a quadprog backend, integrating the IK velocity for up to five iterations.

\paragraph{Evaluation Settings}
We evaluate each solved configuration by applying forward kinematics and measuring the $\ell_2$ distance between the reconstructed keypoints and the predicted keypoints. On the test set of Dexonomy Benchmark, we solve a total of $128{,}340$ IK attempts, corresponding to $2{,}139$ objects $\times$ 6 scales $\times$ 10 grasp proposals.

\begin{figure}[h]
    \centering
    \vspace{-2mm}
    \includegraphics[width=0.8\linewidth]{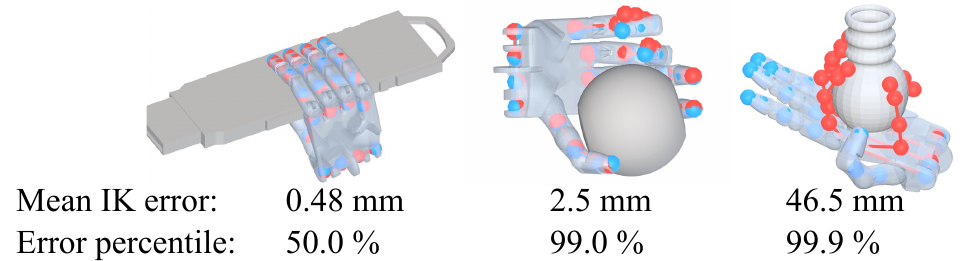}
    \caption{Predicted keypoints (red) and FK keypoints reconstructed from the IK solution(blue). The third example shows a rare backward-folding failure ($\sim 0.2\% $ of grasps).}
    \label{fig:IK_error}
\end{figure}

\paragraph{Results}
Table~\ref{tab:ik_analysis} reports the IK accuracy and runtime. Overall, 97.37\% of the solutions achieve a maximum keypoint reconstruction error below 5 mm. We also observe that grasps with larger keypoint reconstruction errors are more likely to fail in downstream evaluation. The average mean keypoint error is only 0.669 mm, and the average maximum keypoint error is 1.861 mm.  As shown in Fig. ~\ref{fig:IK_error}, among 128k test grasps, 99\% have a mean IK reconstruction error below 2.5 mm, and 50\% below 0.5 mm. Only approximately 0.2\% exhibit clear failures, typically with fingers folding backward toward the dorsum. Explicit feasibility constraints can be imposed on the estimated terminal state at intermediate flow times, but doing so provides only marginal gains: the failure rate decreases from 0.2\% to 0.08\%, while GSR improves by only 0.4 percentage points.

These results show that the network predictions are already highly compatible with hand kinematics, and predicted keypoints can be reliably projected back to configuration space by IK, supporting their use as an intermediate representation. The runtime overhead is also small: the amortized runtime with 64 workers on a 32-core AMD EPYC CPU is approximately 0.002 s per grasp.

\begin{table}[h]
    \centering
    \small
    \begin{tabular}{lc}
        \toprule
        Metric & Value \\
        \midrule
        IK max keypoint error $\leq$ 5 mm & 97.37\% \\
        Avg. mean keypoint error & 0.669 mm \\
        Avg. max keypoint error & 1.861 mm \\
        64-worker amortized runtime & 0.002 s \\
        \bottomrule
    \end{tabular}
    \vspace{8pt}
    \caption{\textbf{IK projection accuracy and runtime.} Predicted keypoints can be accurately and efficiently projected into configuration space.}
    \label{tab:ik_analysis}
\end{table}

\section{Differentiable FK loss for Pose-Joint variant}We trained the pose-joint model using a differentiable FK loss on the estimated terminal
state at intermediate flow times. Although the training loss decreases normally and the generated configurations qualitatively resemble grasps, their quality is poor, yielding only about 1\% GSR. We also added the FK loss to the standard pose-joint flow-matching objective, which achieves 66.0\% GSR,nearly identical to the standard objective alone (66.1\%). Thus, Cartesian keypoint supervision alone does not explain the gain; directly modeling the flow in keypoint space is crucial.

\section{Partial Observation Results} Retraining KPGrasp on single-view point clouds yields 67.4\% GSR, approximately 9\% below the complete-geometry setting. We attribute this gap primarily to the ambiguity of predicting grasps in occluded regions.

\section{Comparison of Different Sampling Steps}

To further improve inference speed, we evaluate whether the number of sampling steps can be reduced. Since a naive 1-step flow matching sampler yields a near zero success rate, we adopt a distilled FlowMap~\cite{boffi2025build} variant for 1-step generation and use our flow-matching model for 3-step, 5-step and 7-step generation with Euler integration. All variants are evaluated for end-to-end inference and the likelihood is computed with discretized trace integration as in the main paper. For multistep variants, we accumulate trace estimates along the sampled ODE trajectory; For 1-step variants, we approximate the trace integral over $[0,1]$ using the average velocity.
The 1-step FlowMap variant is implemented with the Lagrangian Self-Distillation (LSD) objective of~\cite{boffi2025build}. Specifically, the model is conditioned on the current time $t$ and target time $r$ to predict the transition from $x_t$ to $x_r$. At inference time, setting the solver to one step corresponds to $t=0$ and $r=1$, allowing the model to generate a grasp with a single network forward pass.

The 1-step variant achieves the fastest inference time of 0.01 s, but its success rate drops to 52.7\% even with FlowMap distillation. Although the 3-step variant is faster than the 5-step variant, it obtains a lower success rate; using more than five steps does not further improve performance. We therefore use 5 Euler steps in the final model, as they provide the best trade-off between grasp quality and inference speed.

\begin{table}[h]
    \centering

    \begin{tabular}{lcc}
        \toprule
        Method & Success Rate (\%) & Inference Time (s) \\
        \midrule
        %DexGrasp-Anything & 21.8 & 2.8 \\
        %Dexonomy(w/classifier) & 63.9 & 0.035 \\
        Ours (1-step FlowMap) & 52.7 & \textbf{0.01} \\
        Ours (3-step flow matching) & 66.5 & 0.024 \\
        Ours (5-step flow matching) & \textbf{76.3} & 0.032 \\
        Ours (7-step flow matching) & 76.0 & 0.04 \\
        \bottomrule
    \end{tabular}
    \vspace{8pt}
    \caption{\textbf{Comparison of different sampling steps.} We use 5 Euler steps in the final model because they provide the best trade-off between quality and speed.}
    \label{tab:inference_time}
\end{table}

\section{Additional Results of DexGrasp Anything Benchmark}

% \subsection{Training on DexGrasp Anything Datasets}
\label{sec:dga_training}

As mentioned in the main paper, we further evaluate a setting in which KPGrasp is trained directly on each benchmark dataset. As shown in Table~\ref{tab:DGAbenchmark}, KPGrasp trained on each corresponding dataset achieves a lower average success rate than DexGrasp Anything model. We attribute this to two possible factors. First, the original training datasets achieve only a 42.8\% average success rate when directly evaluated under the DexGrasp Anything (DGA) benchmark protocol, likely because of a mismatch between the DGA testing metric and the dataset filtering protocol. Second, KPGrasp learns directly from the data distribution and is therefore sensitive to training data quality, whereas DexGrasp Anything incorporates hand-crafted priors that are less sensitive to the training data. In contrast, KPGrasp trained on Dexonomy achieves the best average success rate of 63.8\%, indicating that a large, high-quality grasp dataset can provide a more transferable grasp prior.

\begin{table}[h]
    \centering
    \begin{tabular}{lcccccc}
    \toprule
         & GRAB & RealD & MD & UDG & DGN & Avg. \\
         \midrule
        Original DGA Training Datasets & \textbf{62.8} & 32.7 & 40.6 & 22.8 & 55.2 & 42.8\\
        DGA Model & 57.9 & 44.8 & \textbf{79.1} & \underline{53.1} & 57.5 & \underline{58.5}\\
        KPGrasp (Trained on each dataset) & 44.7 & \underline{45.6} & 54.7 & 33.1 & \underline{61.8} & 47.8 \\
        KPGrasp (Trained on Dexonomy) & \underline{59.8} & \textbf{67.1} & \underline{62.2} & \textbf{62.3} & \textbf{67.5} & \textbf{63.8} \\
        \bottomrule
    \end{tabular}
    \vspace{8pt}
    \caption{\textbf{Success rate (\%) on the DexGrasp Anything benchmark.}
         \textbf{Bold} indicates the best performance, and \underline{underlined} indicates the second best. We analyze the results in Sec.~\ref{sec:dga_training}.
    }
    \label{tab:DGAbenchmark}
\end{table}

\section{Additional Details of Real World Experiment}

\subsection{Sharpa Hand Keypoint Annotation}
\begin{figure}[h]
    \centering
    \vspace{-2mm}
    \includegraphics[width=0.4\linewidth]{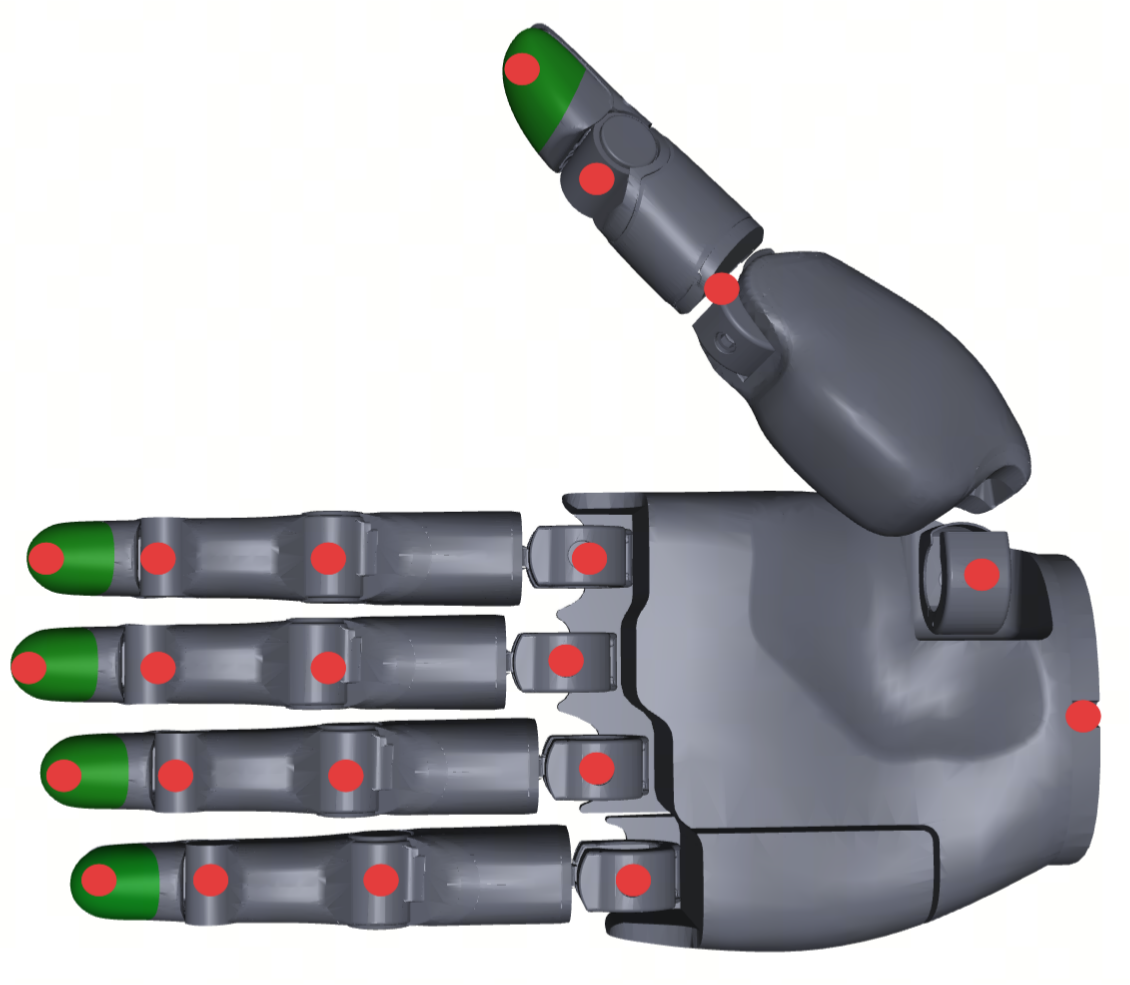}
    \caption{\textbf{Sharpa Hand Keypoint Annotation.}
    We annotate $J=21$ keypoints on Sharpa Hand for real world experiment.}
    \label{fig:sharpa_kp}
\end{figure}

We use $J=21$ keypoints for Sharpa Hand as shown in Fig.~\ref{fig:sharpa_kp}. Keypoints are placed at the mechanical joints and fingertips, with five additional keypoints on the rigid palm.

\subsection{Implementation Details}

To generate training data, we first retarget the Shadow Hand grasp templates in Dexonomy to the Sharpa Hand and then regenerate tabletop grasps with the Dexonomy analytic pipeline, producing 3.3M grasps over 5k DexGraspNet (DGN) objects~\cite{wang2022dexgraspnet}. We render partial point clouds by placing the camera at a random height of 30--40 cm and at a horizontal distance of 30--50 cm from the object. We retrain KPGrasp on this dataset.  

At deployment, we segment the target object point cloud in the robot base frame using GroundedSAM~\cite{ren2024grounded}, estimate the table plane to align the observation with the simulated partial point cloud distribution, generate 100 grasp candidates, and rank them by likelihood. We then perform trajectory interpolation and filter candidates that introduce collisions. We observe three main collision modes: (i) the generated hand pose is very close to the table plane; (ii) The unused camera mounted on the right wrist may collide with the table after solving IK for the right arm; and (iii) when the initial hand pose and target hand pose lie on opposite sides of the object, the interpolated trajectory may collide with the object. Future work could address these issues with collision-aware motion planning, such as cuRobo, and by leaving a larger hand-table margin during simulation data generation.

% After filtering these collision modes, the generated grasps can be executed reliably. The resulting grasps achieve a high success rate across diverse objects and geometries while retaining the diversity observed in simulation.

% \section{Tokenization or Parameterization?}

% In this section, we compare different parameterizations and show that the improvement mainly comes from the output parameterization rather than the tokenization scheme.

% \begin{table}[h]
%     \caption{\textbf{Ablation Study on Hand Parameterization and Tokenization.} Improvement mainly comes from keypoint parametrization.}
%     \centering
%     \resizebox{\linewidth}{!}{
%     \begin{tabular}{lccccc}
%     \toprule
%     Method & GSR($\%$)$\uparrow$ & OSR($\%$)$\uparrow$& CDC($mm$)$\downarrow$& PD($mm$)$\downarrow$ & Diversity$\downarrow$ \\
%     \midrule
%         Pose-Joint as 1 token & 62.5 & 92.0 & 10.3 & 3.7 & 25.7  \\
%         Pose-Joint as 34 tokens & 66.1 & 95.1 & 9.5 & 3.2 & 25.7\\
%         Pose-Joint as 24 tokens & 63.5 & 94.6 & 3.2 & 3.3 & 25.8 \\
%         \midrule
%         Keypoint as 1 token & 55.4 & 93.2 & 10.1 & 8.6 & 23.7 \\
%         Keypoint as $J=21$ tokens (Ours) & \textbf{76.3} & \textbf{97.3} & \textbf{5.6} & \textbf{2.4} & \textbf{23.3} \\
%     \bottomrule
%     \end{tabular}
%     }
%     \label{tab: param ablation}
%     \vspace{-3mm}
% \end{table}

\end{document}